\documentclass[a4paper,twocolumn]{article}
\usepackage[T1]{fontenc}
\usepackage[utf8]{inputenc}
\usepackage{times}
\usepackage{SCITEPRESS}
\usepackage{graphicx,amsmath,amssymb,booktabs,tabularx,array}
\usepackage{apalike}
\usepackage{microtype,etoolbox,placeins,flushend}
\makeatletter
\def\title#1{\gdef\@title{#1}}
\patchcmd{\maketitle}{\vskip -0.07in}{\vskip 0.15in}{}{}
\patchcmd{\maketitle}{\vskip 0.51in}{\vskip 0.15in}{}{}
\makeatother
\usepackage[hidelinks]{hyperref}
\hypersetup{pdftitle={Observer Choice and Threshold Selection in Retinal Vessel Segmentation},pdfauthor={Wenhao Xu, Yixian Kong, Ting Pan, Feilong Wang, Changwei Wang},pdfsubject={Research manuscript},pdfkeywords={retinal vessel segmentation, inter-observer variation, threshold selection}}
\graphicspath{{figures/}}
\newcommand{\ainote}{\par\noindent{\footnotesize\emph{Drafting assistance:} \cite{tool}; see disclosure.}\par}
\newcommand{\DiceMetric}{\mathrm{Dice}}
\newcommand{\ThresholdGrid}{\mathcal{T}}
\newcommand{\ValidationSet}{\mathcal{V}}
\newcommand{\HumanDice}{77.65}
\newcommand{\HumanDisagreement}{3.05}
\newcommand{\HumanUnionDisagreement}{36.47}
\newcommand{\VesselFractionOne}{6.93}
\newcommand{\VesselFractionTwo}{6.64}
\newcommand{\RFBaseWorst}{70.53}
\newcommand{\RFBaseOne}{73.66}
\newcommand{\RFBaseTwo}{71.06}
\newcommand{\RFBasePrecision}{72.23}
\newcommand{\RFBaseRecall}{76.27}
\newcommand{\RFBaseDisagreement}{48.47}
\newcommand{\RFMinWorst}{70.45}

\newcommand{\RFMinPrecision}{73.09}
\newcommand{\RFMinRecall}{75.14}
\newcommand{\RFMinDisagreement}{47.00}

\newcommand{\RFDelta}{-0.073}
\newcommand{\RFDeltaLow}{-0.384}
\newcommand{\RFDeltaHigh}{0.238}
\newcommand{\RFShiftCount}{19}
\newcommand{\RFTrainSeconds}{4.44}
\newcommand{\RFPredictSeconds}{0.79}
\newcommand{\RFTuningGain}{12.31}
\newcommand{\RFAPOne}{0.8109}
\newcommand{\RFAPTwo}{0.7740}
\newcommand{\RFAUCOne}{0.9764}
\newcommand{\RFAUCTwo}{0.9732}
\newcommand{\ETBaseWorst}{68.80}
\newcommand{\ETBaseOne}{71.96}
\newcommand{\ETBaseTwo}{69.33}

\newcommand{\ETMinWorst}{68.68}

\newcommand{\ETDelta}{-0.119}
\newcommand{\ETDeltaLow}{-0.445}
\newcommand{\ETDeltaHigh}{0.168}
\newcommand{\ETShiftCount}{16}
\newcommand{\ETTrainSeconds}{0.66}
\newcommand{\ETPredictSeconds}{0.84}
\newcommand{\ETTuningGain}{10.77}
\newcommand{\ETAPOne}{0.7937}
\newcommand{\ETAPTwo}{0.7501}
\newcommand{\ETAUCOne}{0.9742}
\newcommand{\ETAUCTwo}{0.9708}

\begin{document}
\title{Observer Choice and Threshold Selection in Retinal Vessel Segmentation: A Subject-Separated Evaluation}
\author{\authorname{Wenhao Xu$^{1}$, Yixian Kong$^{2}$, Ting Pan$^{2}$, Feilong Wang$^{2}$, Changwei Wang$^{3,*}$}
\affiliation{$^{1}$Zhengzhou Police University, Zhengzhou, China}
\affiliation{$^{2}$School of Artificial Intelligence, Beijing University of Posts and Telecommunications, Beijing, China}
\affiliation{$^{3}$Qilu University of Technology and Jinan Supercomputing Center, Jinan, China}
\email{changweiwang@sdas.org; *Corresponding author: Changwei Wang}}
\keywords{Retinal Vessel Segmentation, Multiple Annotations, Threshold Selection, Subject-Level Evaluation.}
\abstract{The annotation used to select a segmentation threshold is part of the evaluation protocol, yet its effect is easily conflated with model quality. We examine this choice for retinal vessel segmentation using all 28 CHASE\_DB1 images and both human annotations. A fixed seven-fold protocol keeps both eyes of each of the 14 subjects together. Random forests and Extra Trees are fitted against observer 1 with three random seeds, yielding 42 fits. Five threshold policies share identical score maps: fixed 0.50, observer-1 tuning, observer-2 tuning, mean-observer tuning, and maximin tuning of the per-image lower observer Dice. For random forests, maximin changes the threshold in \RFShiftCount{} of 21 fits, but worst-observer Dice decreases from \RFBaseWorst\% to \RFMinWorst\%. The paired difference is \RFDelta{} percentage points, with a conditional subject-bootstrap 95\% interval of [\RFDeltaLow, \RFDeltaHigh]. Extra Trees shows the same direction. Identical observer-1-tuned random-forest masks score \RFBaseOne\% against observer 1 and \RFBaseTwo\% against observer 2. The results support explicit reporting of both the threshold-selection reference and evaluation reference; they do not support an accuracy benefit from maximin tuning in this cohort. All splits, raw predictions, metrics and code are supplied. AI assistance is disclosed \cite{tool}.}

\maketitle

\section{INTRODUCTION}
\label{sec:intro}
A retinal vessel segmentation score measures agreement with a particular reference annotation. It does not establish agreement with every plausible tracing of the same image. Public datasets such as DRIVE, STARE and CHASE\_DB1 have made vessel extraction reproducible \cite{staal2004,hoover2000,fraz2012}, but the existence of two human annotations creates an additional evaluation choice: which observer supplies the validation objective, and which supplies the reported test score?

This choice matters even when the segmentation model is held fixed. A model usually produces a continuous vessel score, while overlap measures require a binary mask. Selecting a threshold against one observer can favor that observer's annotation convention. Evaluating only against the same convention leaves its transfer to the other observer unmeasured. Reporting two test scores is useful, but a complete description also needs the annotation used for threshold selection. The objective used to select an operating point and the reference used to evaluate it are separate parts of the experiment.

We study this issue on CHASE\_DB1 using both eyes of all 14 subjects and both supplied vessel annotations. The experiment compares five global threshold policies on identical out-of-fold score maps: a fixed threshold, selection against either observer individually, selection by their mean Dice, and selection by the mean per-image lower Dice. We refer to the last policy as \emph{maximin}. The experiment is repeated with random forests and extremely randomized trees using three random seeds. All choices of features, splits, thresholds and primary outcome were fixed locally before fitting.

The contribution is a controlled evaluation of observer-dependent operating points. It comprises a subject-separated protocol, a paired comparison that isolates threshold selection from model fitting, and a complete record of every fit and held-out image. The classifiers, image features and scalar threshold search are established tools; they are not presented as a new vessel segmentation architecture. The experiment also does not assume that optimizing the lower validation score must improve its held-out counterpart. That distinction is central when only two subjects are available for threshold selection.
\ainote

\section{RELATED WORK}
\label{sec:related}
\subsection{Retinal Vessel Segmentation}
Classical vessel extraction uses image structure at several spatial scales. Matched-filter threshold probing \cite{hoover2000}, ridge-based features \cite{staal2004}, Hessian analysis \cite{frangi1998}, and multiscale line responses \cite{nguyen2013} provide complementary ways to describe elongated structures. Fraz et al.\ combine vessel descriptors with bagged and boosted decision trees \cite{fraz2012}. These works motivate a compact multiscale baseline, although the feature set and training procedure used here do not reproduce any of those complete systems.

Deep models address spatial context and vessel continuity more directly. U-Net introduced an encoder--decoder architecture with connections between corresponding resolutions \cite{ronneberger2015}. DRIU specializes convolutional features for vessel and optic disc segmentation \cite{maninis2016}. IterNet refines segmentations using repeated small U-Nets \cite{li2020}, while DA-Net combines local and global information with adaptive strip upsampling \cite{wang2022}. Study Group Learning addresses incomplete vessel labels through learned supervision \cite{zhou2021}. These approaches concern representation or learning; the present comparison concerns the operating point of an already fitted model. Their reported scores are therefore not treated as comparable experimental baselines.

\subsection{Multiple References and Evaluation}
Several methods explicitly address disagreement between annotators. STAPLE estimates a probabilistic reference and the performance of its contributing segmentations \cite{warfield2004}. The probabilistic U-Net models multiple plausible outputs \cite{kohl2018}. The diagnosis-first framework of Wu et al.\ uses diagnostic performance to guide multi-observer label fusion \cite{wu2022}. Our experiment retains both annotations as separate references and does not estimate which observer is more reliable.

Threshold optimization for F1 is already well studied \cite{lipton2014}. Medical segmentation evaluation also requires attention to the choice and aggregation of metrics \cite{taha2015,maier2024,reinke2024}. A recent retinal segmentation preprint considers how uncertainty estimates support deferral decisions \cite{maganti2026}; here the decision is the binary vessel mask itself, with no deferral or clinician intervention. We use \emph{threshold selection} rather than \emph{probability calibration}: changing a decision threshold does not make a score a calibrated probability. The specific question is whether an observer-balanced validation objective transfers to unseen subjects under a fixed, reproducible segmentation pipeline.
\ainote

\section{METHODS}
\label{sec:methods}
\subsection{Score Maps and Image Features}
Each native-resolution photograph is mapped to 29 pixel features. Five are the three RGB intensities divided by 255, a green-to-red ratio with a $1/255$ denominator stabilizer, and a broad background-minus-green contrast with Gaussian scale 32 pixels. The green image is independently normalized using its first and 99th intensity percentiles and clipped to $[0,1]$. Percentiles are calculated inside the largest filled connected component where the largest RGB channel exceeds 10 on the original 8-bit scale. This intensity-only region is used for normalization, never to mask the evaluation.

For each $\sigma\in\{1,2,4,8\}$ pixels, six further features are calculated from normalized green: Gaussian-smoothed intensity, smoothed-minus-original contrast, $\sigma$-normalized gradient magnitude, the two algebraically ordered eigenvalues of the $\sigma^2$-normalized Hessian, and local standard deviation. Thus the feature count is $5+4\times6=29$. Features retain the original $999\times960$ resolution. No image augmentation, learned enhancement, component removal or morphological postprocessing is applied.

The two estimators are a random forest (RF) \cite{breiman2001} and Extra Trees (ET) \cite{geurts2006}, implemented in scikit-learn 1.8.0 \cite{pedregosa2011}. Both use 64 trees, maximum depth 18, minimum leaf size 10 and square-root feature subsampling. RF uses bootstrap samples and ET uses the full sampled training set. For each training image, 2,000 observer-1 vessel pixels and 2,000 background pixels are sampled without replacement. Twenty training images yield 80,000 pixels per fit. Classifier scores are averages of leaf class proportions. Because training is class balanced, we do not interpret these scores as clinical probabilities.

\subsection{Observer-Dependent Thresholds}
Let $p_i(x)$ be the vessel score for pixel $x$ of image $i$, and $y_{io}$ the annotation from observer $o\in\{1,2\}$. At threshold $\tau$, the prediction is
\begin{equation}
 \hat y_i^\tau(x)=\mathbf{1}\{p_i(x)\geq\tau\}.
\end{equation}
With $D_{io}(\tau)=\DiceMetric(\hat y_i^\tau,y_{io})$, the three validation objectives are
\begin{align}
 J_o(\tau)&=\frac{1}{|\ValidationSet|}\sum_{i\in\ValidationSet}D_{io}(\tau),\\
 J_{\mathrm{mean}}(\tau)&=\frac{1}{|\ValidationSet|}\sum_{i\in\ValidationSet}
       \frac{D_{i1}(\tau)+D_{i2}(\tau)}{2},\\
 J_{\mathrm{min}}(\tau)&=\frac{1}{|\ValidationSet|}\sum_{i\in\ValidationSet}
       \min_{o\in\{1,2\}}D_{io}(\tau).
\end{align}
The O1-tuned and O2-tuned policies maximize $J_1$ and $J_2$, respectively. Mean-tuned maximizes $J_{\mathrm{mean}}$, and maximin maximizes $J_{\mathrm{min}}$. Each searches the same 91 thresholds,
$\ThresholdGrid=\{0.05,0.06,\ldots,0.95\}$, choosing the smallest in an exact tie. Fixed 0.50 provides an untuned reference. One threshold is selected per fitted model and applied to every test image assigned to that model.

The minimum in $J_{\mathrm{min}}$ is taken \emph{before} averaging images. In general this differs from the smaller of the two observer-average Dice scores. It allows the limiting observer to vary across images. It also gives neither observer a privileged role in threshold selection, although observer 1 remains the sole training reference. The five policies use exactly the same feature maps and classifier outputs within each fit.

\subsection{Outcomes and Aggregation}
For each observer, Dice and IoU are computed from true positives (TP), false positives (FP) and false negatives (FN):
\begin{align}
 \DiceMetric&=\frac{2\mathrm{TP}}{2\mathrm{TP}+\mathrm{FP}+\mathrm{FN}},\\
 \mathrm{IoU}&=\frac{\mathrm{TP}}{\mathrm{TP}+\mathrm{FP}+\mathrm{FN}}.
\end{align}
Precision and recall are also retained. The primary outcome $W$ is the mean of the lower observer Dice for each image, first averaged over the two eyes and three seeds within each subject, and then over the 14 subjects. Averaging scores across seeds does not ensemble their predictions. The prespecified primary contrast is RF maximin minus RF O1-tuned in $W$; ET provides a second estimator comparison on the same cohort.

Average precision (AP) and ROC AUC are calculated once per held-out score map and observer, then averaged across images and seeds. AP complements ROC AUC in this imbalanced setting \cite{saito2015}. These measures cannot distinguish threshold policies sharing a score map. We additionally record the fraction of predicted vessel pixels in the disagreement region $y_{i1}\mathbin{\triangle}y_{i2}$, without interpreting agreement with either observer as an adjudicated truth.
\ainote

\section{EXPERIMENTAL PROTOCOL}
\label{sec:protocol}
\subsection{Data and Subject Separation}
CHASE\_DB1 contains 28 photographs from both eyes of 14 children, with two independent manual vessel annotations per photograph \cite{fraz2012}. We use the complete distributed set at $999\times960$ pixels and evaluate every image pixel. Images are not resized and no annotation-derived field-of-view mask is applied. These choices must accompany the scores: our results do not share the split or evaluation region of every published CHASE\_DB1 experiment.

\begin{figure*}[t]
 \centering
 \includegraphics[width=\textwidth]{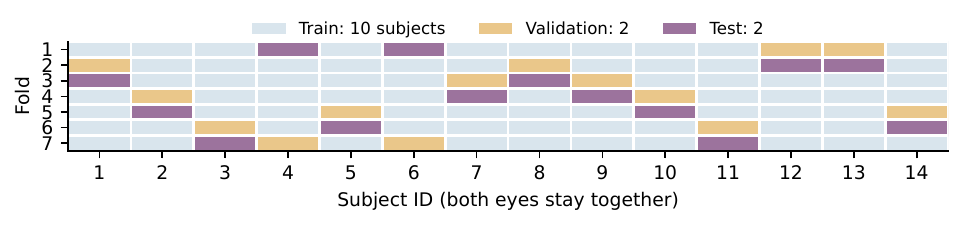}
 \caption{The fixed seven-fold allocation. Each column represents a subject and both eyes. The four validation images select the threshold; the four test images are scored only after that choice is recorded.}
 \label{fig:splits}
\end{figure*}

The 14 subjects are permuted with NumPy random seed 20260909 and divided into seven groups of two. In fold $k$, group $k$ is tested, group $(k+1)\bmod7$ selects the threshold, and the remaining ten subjects train the model. Each subject is tested once; both eyes always remain in the same partition. Figure~\ref{fig:splits} gives the actual allocation. The same partitions are used by all estimators and seeds.

\subsection{Repetition, Uncertainty and Audit}
Seeds 17, 41 and 83 vary pixel sampling and estimator randomness, producing $2\times3\times7=42$ fitted models and 168 held-out score maps. The protocol, split manifest, data manifest and experiment-source hashes were frozen locally before fitting. This is a documented analysis-plan freeze, not an externally registered study. Thresholds are written to a separate record before test annotations are used for scoring. No hyperparameters or comparison policies were revised after test prediction.

Paired percentile intervals use 10,000 bootstrap resamples of the 14 subject-level values, with seed 20260910. Eyes and technical seeds are averaged within a subject before resampling. The intervals condition on the stored out-of-fold predictions. They do not capture uncertainty from retraining on newly sampled cohorts, and overlapping training folds limit an independent-sample interpretation. We report no pixel-level significance tests and do not treat the three seeds as additional patients.

All raw confusion counts, continuous score maps, validation curves, selected thresholds, runtime records and environment versions accompany the source. Exact threshold equality, confusion counts, Dice and IoU were checked against independent scikit-learn metric calculations. Qualitative images 01L, 07L and 14L were specified before fitting; their RF seed-17 predictions are displayed with a common fixed crop.
\ainote

\section{RESULTS}
\label{sec:results}
\subsection{Agreement and Primary Outcome}
The mean human--human Dice is \HumanDice\%. Observer 1 labels \VesselFractionOne\% of image pixels as vessels and observer 2 labels \VesselFractionTwo\%. Their disagreement covers \HumanDisagreement\% of all pixels, or \HumanUnionDisagreement\% of the vessel union when calculated per image and then averaged. Human agreement is descriptive; it is not used as an upper bound on algorithm performance.

\begin{figure*}[t]
\centering
\includegraphics[width=\textwidth]{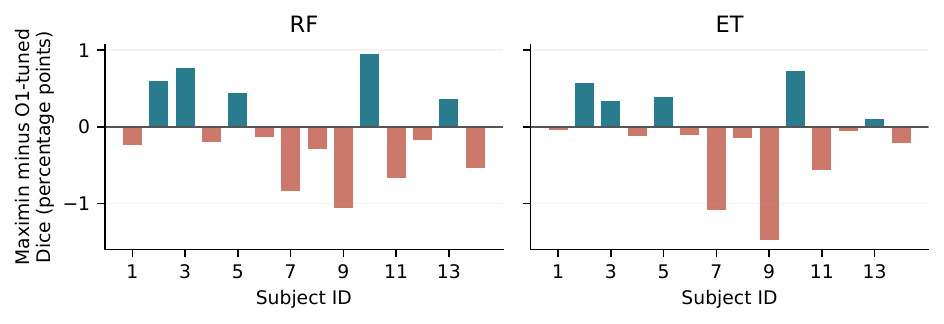}
\caption{Every subject's change in worst-observer Dice, after averaging both eyes and three seeds. Positive values favor maximin; negative values favor O1-tuned. Subject IDs follow the public data filenames.}
\label{fig:paired}
\end{figure*}

Table~\ref{tab:main} reports all policies. RF maximin obtains $W=\RFMinWorst\%$, compared with \RFBaseWorst\% for O1-tuned. The prespecified paired contrast is \RFDelta{} percentage points (conditional 95\% interval [\RFDeltaLow, \RFDeltaHigh]). ET obtains \ETMinWorst\% and \ETBaseWorst\%, respectively, with a contrast of \ETDelta{} points [\ETDeltaLow, \ETDeltaHigh]. Both intervals span zero. Maximin improves five subjects and reduces the score for nine subjects with either estimator (Figure~\ref{fig:paired}). Its mean contrast is negative in all three seeds for both estimators. These observations do not support the hypothesized held-out improvement.

\begin{table*}[t]
\centering
\caption{Native-resolution, whole-image evaluation. Values are percentages, averaged over eyes and seeds within subjects. $W$ averages the lower observer Dice per image; it is not the smaller of the two displayed Dice columns. Intervals condition on fitted predictions.}
\label{tab:main}
\small
\setlength{\tabcolsep}{9pt}
\begin{tabular}{llrrrr}
\toprule
Estimator & Threshold policy & Dice O1 & Dice O2 & $W$ & 95\% interval for $W$\\
\midrule
\csname @@input\endcsname results_table.tex
\bottomrule
\end{tabular}
\end{table*}

\subsection{Threshold Changes and Reference Choice}
Maximin selects a different threshold from O1-tuned in \RFShiftCount{} of 21 RF fits and \ETShiftCount{} of 21 ET fits. Per-fit thresholds and validation curves are provided in the research package. RF maximin thresholds range from 0.78 to 0.90; ET thresholds range from 0.71 to 0.85. Frequent changes in the operating point therefore do not imply a beneficial change in held-out overlap.

\begin{figure*}[t]
\centering
\includegraphics[width=\textwidth]{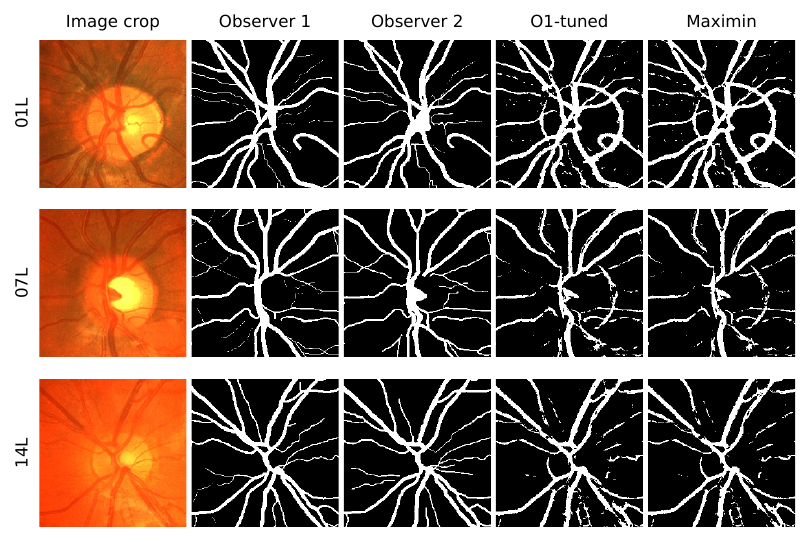}
\caption{Predetermined RF seed-17 examples. Each panel shows the same $400\times400$ crop, rows 230--629 and columns 290--689 in zero-based coordinates. The original photographs are cropped only for display; masks are the supplied annotations or actual held-out predictions. Image credit: CHASE\_DB1 creators \cite{fraz2012}, CC BY 4.0. No image synthesis or generative editing is used.}
\label{fig:qual}
\end{figure*}

For RF, O1-tuned and maximin label \RFBaseDisagreement\% and \RFMinDisagreement\% of disagreement-region pixels as vessels. Against observer 1, precision changes from \RFBasePrecision\% to \RFMinPrecision\%, while recall changes from \RFBaseRecall\% to \RFMinRecall\%. On average, maximin increases precision and reduces recall without improving the primary outcome. Figure~\ref{fig:qual} shows the predetermined cases, including missed small branches and irregular predictions near the bright optic disc. The visual examples illustrate the limited change between threshold policies rather than establish their relative performance.

Reference choice remains consequential after tuning. The same RF O1-tuned masks obtain Dice \RFBaseOne\% against observer 1 and \RFBaseTwo\% against observer 2. ET obtains \ETBaseOne\% and \ETBaseTwo\%. These are paired evaluations of identical predictions, not differences between separately trained observer-specific systems.

\subsection{Untuned Scores and Computation}
O1-tuned exceeds fixed 0.50 by \RFTuningGain{} percentage points in RF $W$, and \ETTuningGain{} points for ET. This is a conventional threshold-selection effect under balanced pixel sampling, not evidence for the maximin policy. AP is \RFAPOne{} against observer 1 and \RFAPTwo{} against observer 2 for RF, versus \ETAPOne{} and \ETAPTwo{} for ET. Corresponding ROC AUC values are \RFAUCOne/\RFAUCTwo{} and \ETAUCOne/\ETAUCTwo. These threshold-independent results apply to every policy of the same estimator.

Mean fitting time is \RFTrainSeconds{} s for RF and \ETTrainSeconds{} s for ET. Prediction from cached features takes \RFPredictSeconds{} s and \ETPredictSeconds{} s per image, respectively, with eight CPU threads on an Intel Xeon Platinum 8370C. Feature extraction takes a further 2.80 s per image. These measurements exclude metric computation and writing the lossless score maps. The final audit recomputed all 840 policy--image records from the 168 stored maps, with zero discrepancy in confusion counts or their derived overlap, precision and recall values.

\ainote

\section{DISCUSSION}
\label{sec:discussion}
\subsection{What the Comparison Establishes}
The principal result is negative: the symmetric maximin validation objective does not improve its held-out counterpart in this experiment. The thresholds often change, but the resulting differences are small and unfavorable on average. A validation objective can increase by construction while its test counterpart decreases. With only four validation images, the selected operating point depends on a small sample of annotation differences. The present design measures that transfer directly, instead of interpreting an optimized validation value as evidence of generalization.

The larger practical issue is reference dependence. The same RF masks receive different Dice scores against the two observers. A reader comparing papers needs both the annotation used to select the threshold and the annotation used for final evaluation. Reporting a second observer only as a human comparator does not describe this selection step. Conversely, reducing the difference between two observer scores would not by itself show a more accurate vessel map: both scores could decline, and neither annotation is an adjudicated biological truth.

This does not make maximin thresholding intrinsically unsuitable. It defines a clear objective when performance against multiple references matters. What is unsupported here is an empirical improvement from that objective under this small validation protocol. Repeating the test with additional independently annotated cohorts and stronger segmentation models would address a different and broader claim.

\subsection{Limitations and Transfer}
The study contains 14 children from one public collection. RF and ET are related tree ensembles sharing the same handcrafted features, not independent replications across model families. No deep network was fitted. Training uses observer 1 throughout, so the experiment is not a symmetric crossover of training observers. The measured effects may change when training against observer 2, fusing labels, increasing the validation cohort, or altering the score distribution.

Whole-image scoring, the native resolution and the balanced training sample also constrain comparisons. Dark background pixels affect ranking metrics, while a 0.50 decision threshold has no special optimality after balanced sampling. The strong improvement over fixed 0.50 should therefore not be advertised as a new algorithmic gain. Other CHASE\_DB1 studies may use different partitions, preprocessing or evaluation regions; their reported numbers cannot be ranked against Table~\ref{tab:main} without a matched rerun.

The bootstrap intervals describe variation across the observed subjects conditional on the fitted predictions. They do not resolve dependence induced by overlapping training folds. Three seeds characterize a limited amount of estimator variability and do not increase the number of independent subjects. We provide the raw paired values so that the size and direction of the effects can be assessed without relying on a significance claim.

Finally, overlap scores do not establish topological correctness, vessel-calibre accuracy, diagnostic benefit or clinical readiness. The qualitative cases show errors that scalar threshold changes cannot repair. Additional expert adjudication and external evaluation would be needed to determine whether a changed mask better represents the vasculature. The present results support a narrower reporting practice: retain both reference-specific scores, state the validation reference explicitly, and keep operating-point selection separate from held-out evaluation.

\ainote

\section{CONCLUSION}
\label{sec:conclusion}
A subject-separated CHASE\_DB1 experiment isolated the annotation used for threshold selection from the fitted segmentation model. Across 42 fits, maximin frequently changed the selected operating point but did not improve worst-observer Dice relative to observer-1 tuning. The mean changes were \RFDelta{} percentage points for RF and \ETDelta{} for ET, with conditional intervals spanning zero. Reference-specific evaluation of identical masks showed a larger numerical difference. These findings support documenting both the threshold-selection annotation and the evaluation annotation, and retaining negative results when a validation objective fails to transfer to unseen subjects. The supplied data, predictions, split manifests and code make the comparison directly auditable.

\ainote

\section*{ACKNOWLEDGEMENTS AND DISCLOSURE}
The public CHASE\_DB1 photographs and annotations are credited to their creators \cite{fraz2012}. The \href{https://researchinnovation.kingston.ac.uk/en/datasets/chasedb1-retinal-vessel-reference-dataset-4/}{institutional data record} supplies CC BY 4.0 terms. A versioned copy was obtained from the public Study Group Learning repository \cite{zhou2021}; that repository's trained models and pseudo-labels were not used. Git blob and SHA-256 checks verify the retrieved mirror files, not an independent byte comparison with the institutional archive.

OpenAI ChatGPT (Codex) \cite{tool} assisted with study planning, source discovery, code development, experiment execution, analysis scripts, the abstract and Sections~1--7, and manuscript revision. Figures are deterministic plots or crops of public images and computed predictions, produced with AI-assisted code; no synthetic patient images or fabricated measurements are used. The human authors retain responsibility for the final submitted content. No AI system is an author. Only publicly released data were analysed; no new participant data were collected.

\bibliographystyle{apalike-compact}
{\small\bibliography{references}}
\end{document}